\documentclass[10pt,twocolumn,letterpaper]{article}

\usepackage[accsupp]{axessibility}

\usepackage[lined,ruled,linesnumbered]{algorithm2e}
\usepackage{animate} 

\usepackage{dsfont}
\usepackage[dvipsnames]{xcolor}
\usepackage[T1]{fontenc}

\usepackage{bmpsize}
\usepackage{epsfig}
\usepackage{float}
\usepackage{graphicx}
\usepackage[justification=raggedright]{caption}
\usepackage{lscape}
\usepackage{subcaption}
\usepackage{wrapfig}

\usepackage{array}
\usepackage{booktabs}                   
\usepackage{colortbl}
\usepackage{makecell}
\usepackage{multirow}
\usepackage{paralist}

\usepackage{amsmath}
\usepackage{pifont}
\usepackage{amssymb}
\usepackage{amsfonts}
\usepackage{bm}
\usepackage{mathtools}

\usepackage{comment}

\colorlet{dark-blue}{blue!50!black}
\colorlet{dark-cyan}{cyan!75!black}
\colorlet{dark-purple}{purple!50!black}
\colorlet{dark-red}{red!75!black}
\colorlet{dark-green}{green!80!black}
\colorlet{dark-orange}{orange!50!black}
\colorlet{dark-gray}{black!75}
\colorlet{light-gray}{black!30}
\definecolor{nice-red}{HTML}{E41A1C}
\definecolor{nice-orange}{HTML}{FF7F00}
\definecolor{nice-yellow}{HTML}{FFC020}
\definecolor{nice-green}{HTML}{39b54a}
\definecolor{nice-blue}{HTML}{0071bc}
\definecolor{nice-purple}{HTML}{984EA3}

\colorlet{verylight-gray}{black!10}
\definecolor{LightCyan}{rgb}{0.66,0.85,0.76}

\makeatletter
\DeclareRobustCommand\onedot{\futurelet\@let@token\@onedot}
\def\@onedot{\ifx\@let@token.\else.\null\fi\xspace}

\makeatother

\newcommand{\ignore}[1]{}   

\newcolumntype{L}[1]{>{\raggedright\let\newline\\\arraybackslash\hspace{0pt}}m{#1}}
\newcolumntype{C}[1]{>{\centering\let\newline\\\arraybackslash\hspace{0pt}}m{#1}}
\newcolumntype{R}[1]{>{\raggedleft\let\newline\\\arraybackslash\hspace{0pt}}m{#1}}

\newcommand{\modelname}{GT-SVJ\xspace}
\newcommand{\modelnamefull}{Generative-Transformer-Based Self-Supervised Video Judge\space}
\newcommand{\backbone}{CogVideoX\xspace}
\newcommand{\backbonewithcite}{CogVideoX~\cite{cogvideox}\xspace}

\definecolor{tabfirst}{rgb}{1, 0.7, 0.7} 
\definecolor{tabsecond}{rgb}{1, 0.85, 0.7} 
\definecolor{tabthird}{rgb}{1, 1, 0.7} 

\graphicspath{{figures}}

\def\mI{\mathcal{I}}

\def\mL{\mathcal{L}}

\def\mN{\mathcal{N}}

\def\mP{\mathcal{P}}
\def\mQ{\mathcal{Q}}

\def\mX{\mathcal{X}}

\def\mZ{\mathcal{Z}}

\def\bbR{\mathbb{R}}

\def\bbE{\mathbb{E}}

\def\1n{\mathbf{1}_n}
\def\0{\mathbf{0}}
\def\1{\mathbf{1}}

\newcommand{\parens}[1]{\left(#1\right)}
\newcommand{\braces}[1]{\left\{#1\right\}}
\newcommand{\bracks}[1]{\left[#1\right]}
\newcommand{\modulus}[1]{\left\vert#1\right\vert}

\usepackage[pagenumbers]{cvpr} 

\definecolor{cvprblue}{rgb}{0.21,0.49,0.74}
\usepackage[pagebackref,breaklinks,colorlinks,allcolors=cvprblue]{hyperref}

\def\paperID{*****} 
\def\confName{CVPR}
\def\confYear{2026}

\title{\modelname: Generative-Transformer-Based Self-Supervised Video Judge For Efficient Video Reward Modeling}

\author{Shivanshu Shekhar \thanks{Work done while an intern at Adobe}\\
University of Illinois Urbana-Champaign\\
{\tt\small shekhar6@illinois.edu}
\and
Uttaran Bhattacharya\\
Adobe Inc.\\
{\tt\small ubhattac@adobe.com}
\and
Raghavendra Addanki\\
Adobe Inc.\\
{\tt\small raddanki@adobe.com}
\and
Mehrab Tanjim\\
Adobe Inc.\\
{\tt\small tanjim@adobe.com}
\and
Somdeb Sarkhel\\
Adobe Inc.\\
{\tt\small sarkhel@adobe.com}
\and
Tong Zhang\\
University of Illinois Urbana-Champaign\\
{\tt\small tozhang@illinois.edu}
}

\begin{document}
\maketitle
\begin{abstract}
Aligning video generative models with human preferences remains challenging: current approaches rely on Vision-Language Models (VLMs) for reward modeling, but these models struggle to capture subtle temporal dynamics. We propose a fundamentally different approach: repurposing video generative models, which are inherently designed to model temporal structure, as reward models. We present the \modelnamefull (\modelname), a novel evaluation model that transforms state-of-the-art video generation models into powerful temporally-aware reward models. Our key insight is that generative models can be reformulated as energy-based models (EBMs) that assign low energy to high-quality videos and high energy to degraded ones, enabling them to discriminate video quality with remarkable precision when trained via contrastive objectives. To prevent the model from exploiting superficial differences between real and generated videos, we design challenging synthetic negative videos through controlled latent-space perturbations: temporal slicing, feature swapping, and frame shuffling, which simulate realistic but subtle visual degradations. This forces the model to learn meaningful spatiotemporal features rather than trivial artifacts. \modelname achieves state-of-the-art performance on GenAI-Bench and MonteBench using only 30K human-annotations: $6\times$ to $65\times$ fewer than existing VLM-based approaches. Project URL:  \url{https://huggingface.co/sasuke-ss1/GT-SVJ}.
\end{abstract}

\section{Introduction}
\label{sec:intro}

\begin{figure}[t]
    \centering
    \includegraphics[width=0.8\columnwidth]{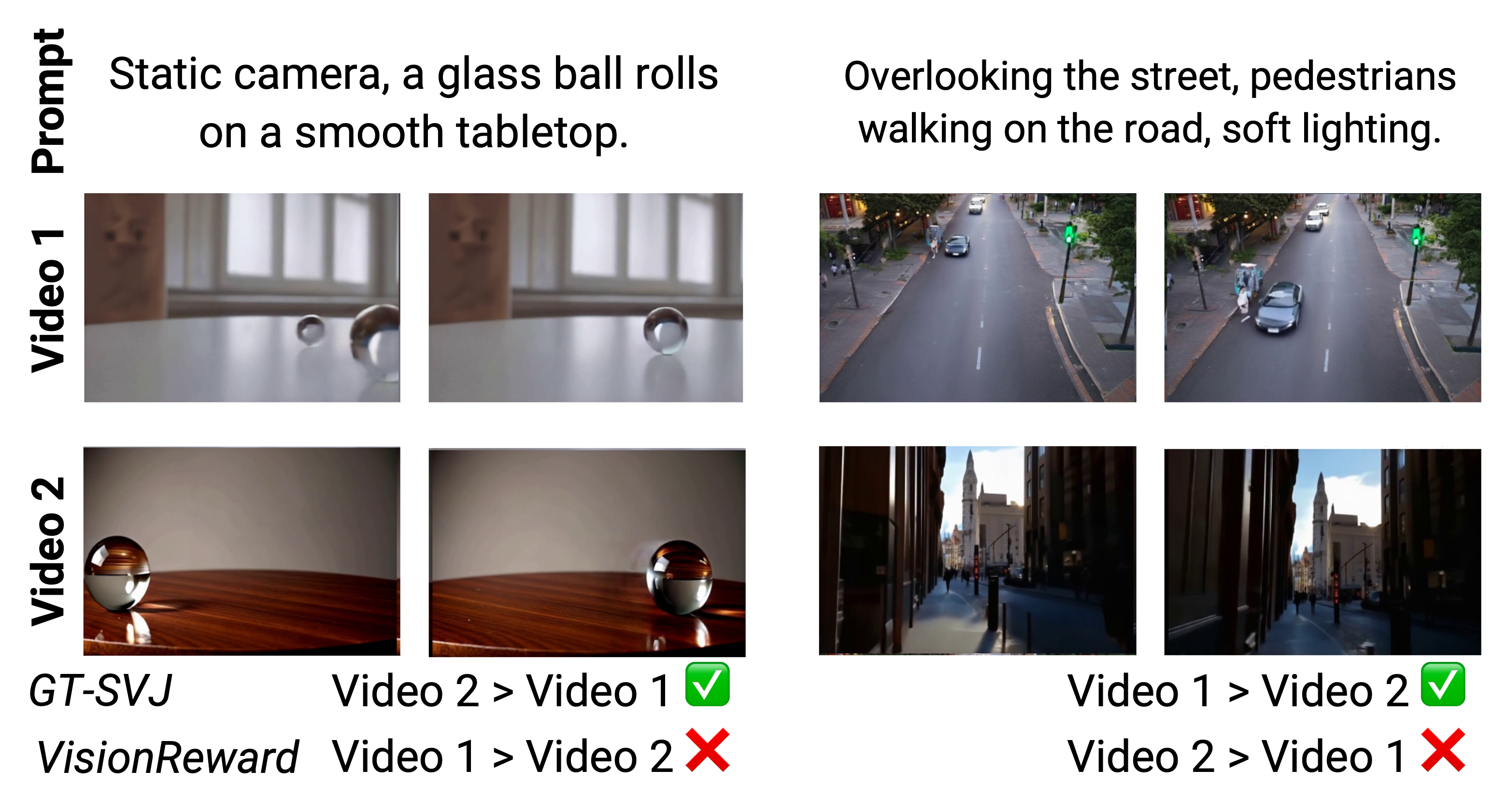}
    \caption{\textbf{\modelname in action.} Given two videos, our self-supervised model evaluates and ranks them on preferences, outperforming baselines on human preference alignment.}
    \label{fig:teaser}
\end{figure}

Recent advancements in video generation have led to highly capable models that can synthesize visually compelling and temporally coherent videos~\cite{PixVerse2024, Adobe2024Firefly, CapCut2024Dreamina, Kuaishou2024KlingAI, Runway2024Gen3, OpenAI2025Sora2, LumaLabs2024DreamMachine, Google2025Veo3}. These models demonstrate impressive fidelity, motion consistency, and scene understanding, marking a significant step toward general-purpose video synthesis. However, despite these achievements, aligning such models with nuanced human preferences remains significantly challenging.

Reinforcement Learning from Human Feedback (RLHF)~\cite{RLHF}, originally introduced to align large language models (LLMs) with human intent, has recently been extended to visual domains such as image and video generation~\cite{DiffDPO, D3PO, SEEDPO, ROCM}. In the learning-from-feedback paradigm, a preference model is trained to capture human preferences of quality, realism, or faithfulness, which is then used to fine-tune the underlying generative model via reinforcement learning or direct preference optimization.

A key obstacle to learning human-preference alignment for videos is the lack of a reliable temporally-aware reward signal. Existing approaches for visual preference modeling typically fine-tune Video-Language Models (VLMs)~\cite{visionreward, videoreward, videoscore} to predict preference scores from pairs of videos. However, these approaches inherently treat videos as collections of independent frames, relying on attention mechanisms to implicitly capture temporal structure. Their frame-centric biases often lead to poor temporal sensitivity to subtle motion quality, temporal smoothness, or consistency across frames. Moreover, these methods rely on large-scale human preference annotations, which are expensive to collect and impede scalability.

To address these limitations, we propose to repurpose a powerful state-of-the-art video generative model (\backbonewithcite) as a temporally grounded reward model for human preferences. Unlike VLMs, modern video generative models are typically trained explicitly to model temporal dependencies through causal self-attention over latent video tokens. By fine-tuning a video generative model as a preference model, we obtain a reward function that consistently captures accurate motion dynamics and temporal coherence, resulting in superior alignment with human preferences. Our results show that generative models, when adapted appropriately, can outperform traditional VLM-based reward models across multiple preference benchmarks while requiring, on average, an order of magnitude less human preference data to match or improve the alignment accuracy with human preferences.

However, naively fine-tuning a generative model for discrimination risks overfitting to limited preference data and learning superficial artifacts over meaningful temporal structure.
To effectively build a preference-based video reward model, we introduce a self-supervised contrastive learning strategy that utilizes energy-based modeling to convert the base video generative model into a discriminative model that can efficiently contrast real and generated videos. To prevent learning trivial domain gaps in the energy landscapes between real and generated videos, we create additional challenging non-real videos by applying controlled perturbations (\textit{e.g.}, frame shuffling, temporal swaps) to the latent representations of real videos, resulting in weak degradations.
The discriminative model, trained on real videos and these weak degradations, learns richer feature representations for subtle spatial and temporal inconsistencies. This leads to a more stable, data-efficient, and meaningful reward signal for downstream preference learning.

In summary, we present a framework and a corresponding model, the \modelnamefull (\modelname), for learning video preferences by combining the temporal understanding of video generative models with the granularity of contrastive self-supervision (\cref{fig:teaser}). Our contributions include:

\begin{itemize}
    \item A self-supervised contrastive learning framework that converts a video generative model to an energy-based discriminative model, improving reward consistency, stability, and data efficiency.
    \item Demonstrating that fine-tuning a video generative model as a reward model yields superior temporal understanding and higher human preference prediction accuracy than existing VLM-based approaches.
    \item Achieving state-of-the-art performance for \modelname on GenAI-Bench~\cite{genai} and MonteBench~\cite{visionreward}, outperforming prior baselines by approximately 25\% and 4\% (with ties) and 3\% and 8\% (without ties) respectively, while remaining competitive on VideoReward-Bench~\cite{videoreward}, trailing the best model by only 4\% (with ties) and 7\% (without ties). Remarkably, our results are achieved using only 30K human-annotated videos: roughly 6$\times$ fewer
  than VideoReward and 65$\times$ fewer than VisionReward, demonstrating exceptional data efficiency.
\end{itemize}


\section{Related Work}
\label{sec:related_work}
Recent text-to-video generation models~\cite{OpenAI2025Sora2, cogvideox,
  Runway2024Gen3, LumaLabs2024DreamMachine} have achieved unprecedented realism,
  making reliable evaluation of motion quality and temporal consistency critical
  for further progress.
  
Evaluating the quality of generated videos remains a fundamental challenge in advancing generative modeling, as it requires metrics that jointly assess spatial fidelity, temporal coherence, and alignment with human perception. Traditional frame-based metrics such as PSNR and SSIM are limited in this regard, since they evaluate individual frames independently and therefore fail to capture the temporal continuity that characterizes realistic videos~\citep{fvd}. To address this, video-level metrics such as the Fr\'echet Video Distance (FVD)~\cite{fvd} and VBench~\cite{vbench} were introduced, leveraging spatiotemporal features from pretrained action recognition models to produce distribution-based measures that correlate more closely with human preferences of overall video quality. Building upon these, the Fr\'echet Video Motion Distance (FVMD)~\cite{fvmd} explicitly models motion by analyzing keypoint trajectories and their velocity and acceleration statistics, providing improved alignment with human assessments, particularly for videos involving complex or rapid movements, where prior metrics like FVD and VBench tend to over-penalize dynamism.

These work highlights the increasing importance of motion-aware evaluation, especially as diffusion-based video generators produce outputs that are both highly realistic and rich in temporal dynamics. Beyond conventional quantitative metrics, evaluation frameworks inspired by natural language understanding, such as Auto-J~\citep{autoj} enable more flexible and context-sensitive assessments through free-form textual feedback. This paradigm is further advanced by LLM-as-a-judge methods~\cite{llmaj}, which incorporate multimodal reasoning to provide both numerical scores and explanatory justifications, making them well-suited for video quality assessment under preference-based settings. Nevertheless, as argued by \cite{reliablejudge}, VLM-based judging systems can exhibit systematic biases or limited content comprehension, potentially undermining their reliability as objective evaluation tools.

Preference-based evaluation provides yet another perspective, incorporating direct human preferences into the evaluation process~\cite{videoreward, videoscore, lift, visionreward}. By leveraging human-in-the-loop comparisons, such methods bridge the gap between automated metrics and subjective preference alignment, allowing video generation models to optimize perceptual and semantic quality simultaneously. More recently, Direct Preference Optimization (DPO)~\cite{dpo} and its diffusion-based variants~\cite{DiffDPO, SEEDPO, D3PO, SPO} have shown that generative models can be fine-tuned directly on preference pairs without the need for an explicit reward function. While these approaches have primarily focused on image generation, their principles extend naturally to video domains, where modeling user preferences over temporal smoothness and motion consistency is equally critical.

Prior efforts have established a broad foundation for video evaluation, spanning frame-level fidelity metrics, spatiotemporal measures, and VLM-based preference scoring. However, VLMs are inherently optimized for video \emph{understanding}, not \emph{generation}, making them ill-suited for capturing the fine-grained temporal structure that differentiates real and synthetic video dynamics; moreover, they often require large amounts of human-annotated data and can exhibit instability or bias. In contrast, our approach leverages a pretrained \emph{generative} video model whose representations inherently capture motion, temporal causality, and fine-grained dynamics. As a result, its learned features provide a stronger and more temporally faithful backbone than those of VLM-based evaluators. Combined with our discriminative contrastive learning using challenging synthetic negatives, the model learns robust temporal cues with significantly less human supervision, yielding a more stable and genuinely video-grounded reward model.


\begin{figure*}[t]
    \centering
    \includegraphics[width=0.8\linewidth]{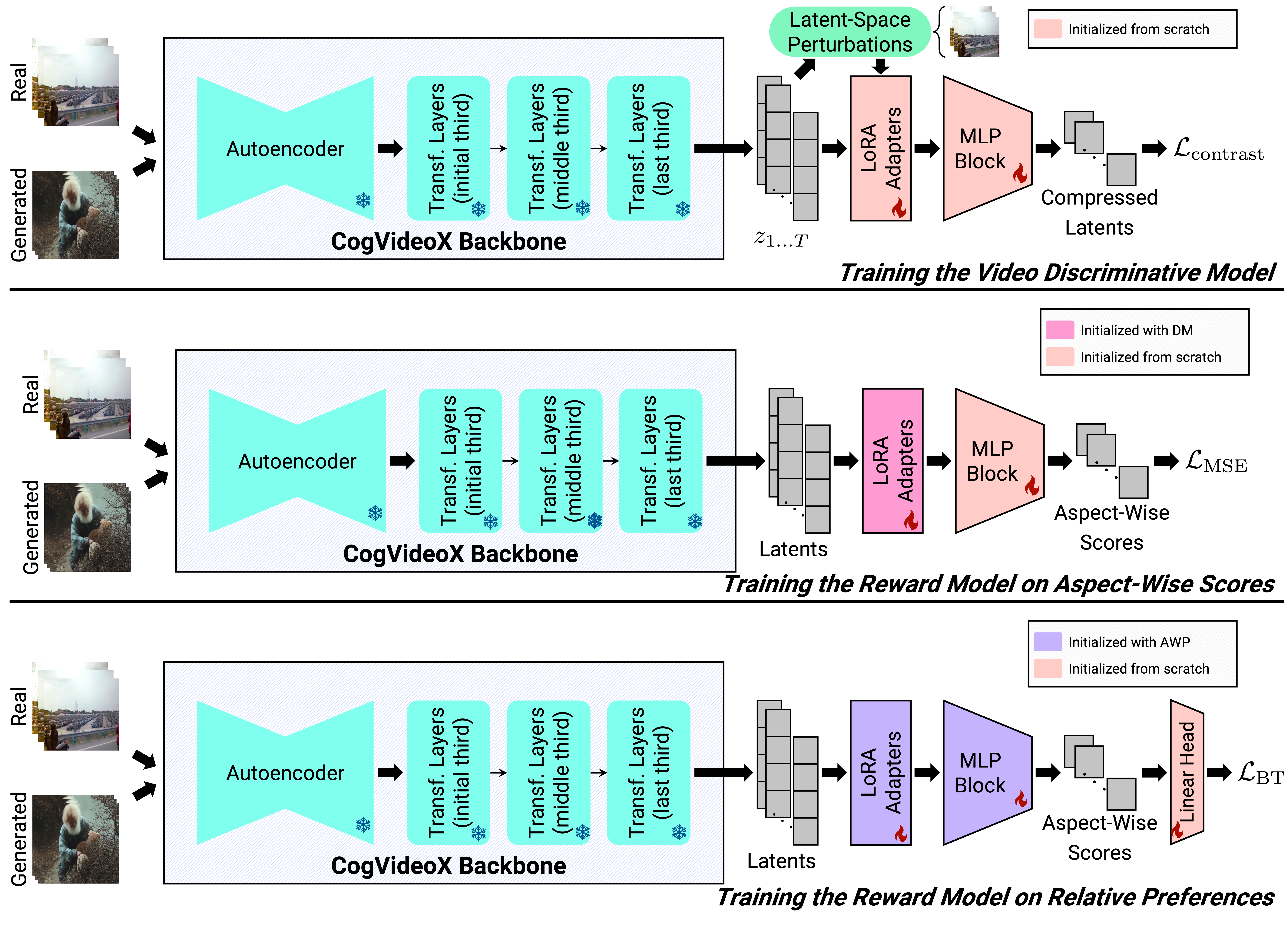}
    \caption{
        \textbf{Overview of the proposed \modelname framework.} The framework consists of two stages: 
        \textit{(top)} \textbf{Training a discriminative model}, where the video generative model (\backbone) is adapted using a contrastive energy-based objective with real, generated, and perturbed videos, and 
        \textit{(middle and bottom)} \textbf{Training a reward model}, where the discriminative model (DM) is aligned with human ratings through aspect-wise prediction (AWP) via regression \textit{(middle)} followed by relative preference modeling \textit{(bottom)}.
    }
    \label{fig:pipeline}
\end{figure*}

\section{Background}
We briefly cover relevant background on energy-based models and reward modeling.

\subsection{Energy-Based Models}

Energy-Based Models (EBMs)~\cite{ebm} define a probability distribution over data samples $x \in \mX$ through an unnormalized energy function $E_\theta : \mX \rightarrow \bbR$ parameterized by $\theta$. The model assigns lower energy to more likely or desirable configurations and higher energy to unlikely ones. Formally, the probability of observing a sample $x$ is expressed as
\begin{equation}
    P_\theta(x) = \frac{\exp(-E_\theta(x))}{Z_\theta},
\end{equation}
where $Z_\theta = \int \exp(-E_\theta(x)) \, dx$ is the partition function for normalization. Since $Z_\theta$ is often intractable, EBMs are typically trained by minimizing an approximate contrastive objective rather than maximizing the exact likelihood \cite{train_energy}.

The contrastive loss encourages the model to assign lower energy to positive samples (real or preferred data) and higher energy to negative samples (non-preferred or perturbed data). A common formulation is
\begin{equation}
\mL_{\text{EBM}} = \bbE_{x^+ \sim p_{\text{data}}} [E_\theta(x^+)] - \bbE_{x^- \sim p_{\text{neg}}} [E_\theta(x^-)],
\label{eq:ebm}
\end{equation}
where $x^+$ denotes samples from the target (real) distribution $p_{\text{data}}$ and $x^-$ denotes samples from the negatively perturbed or generated distribution $p_{\text{neg}}$. In practice, $p_{\text{neg}}$ is often constructed following Langevin Dynamics to stably maintain proximity to the real data~\cite{bayesian_learning_via_ld,l2energy, useless_energy_citation}. A regularization term is usually added \cite{l2energy} to make the training stable, such as:
\begin{equation}
\mL_{2} = \bbE_{x^+ \sim p_{\text{data}}} [E_\theta(x^+)^2] + \bbE_{x^- \sim p_{\text{neg}}} [E_\theta(x^-)^2].
\label{eq:l2}
\end{equation}
Combining \cref{eq:ebm,eq:l2}, the contrastive loss is
\begin{equation}
\label{eq:contrast}
\mL_{\text{contrast}} = \mL_{\text{EBM}} + \beta\mL_{2}.
\end{equation}

In our context, this formulation aligns naturally with reward modeling for generative videos. Here, positive samples correspond to human-preferred or realistic video segments, while negative samples correspond to generated videos and perturbations (\textit{e.g.}, through slicing or feature swapping). Training the reward model under this contrastive objective encourages it to learn a discriminative energy landscape in both spatial and temporal aspects.

\subsection{Reward Modeling}

Reward modeling aims to align model behavior with human preferences by learning a scalar reward function that reflects perceived quality or desirability~\cite{rlhf1, rlhf_useless_citation}. Given a dataset of paired comparisons $\{(x_i^+, x_i^-)\}$, where $x_i^+$ is preferred over $x_i^-$, the goal is to train a function $r_\phi : \mX \rightarrow \bbR$ such that
$r_\phi(x_i^+) > r_\phi(x_i^-)$
for all preference pairs. A probabilistic interpretation of preferences can be derived from the Bradley–Terry (BT) model~\cite{bt}, where the probability that sample $x_i^+$ is preferred over $x_i^-$ is modeled as
\begin{equation}
P(x_i^+ \succ x_i^-) = \frac{\exp(r_\phi(x_i^+))}{\exp(r_\phi(x_i^+)) + \exp(r_\phi(x_i^-))}.
\end{equation}
The reward model is trained to maximize the likelihood of observed preferences, leading to the loss
\begin{equation}
\label{eq:bt}
\mL_{\text{BT}} = - \bbE_{(x^+, x^-)} \bracks{\log P(x^+ \succ x^-)}.
\end{equation}

However, human preferences often include uncertainty or indifference between two samples. The Bradley–Terry with Ties (BTT) model~\cite{btt} extends BT by incorporating an indifference probability controlled by a temperature-like parameter $\gamma$, as
\begin{equation}
\resizebox{0.9\columnwidth}{!}{
    $P(x^+ \succ x^-) = \frac{\exp(r_\phi(x^+)/\gamma)}{\exp(r_\phi(x^+)/\gamma) + \exp(r_\phi(x^-)/\gamma) + 1}$.
}
\label{eq:btt}
\end{equation}
This modification allows the model to better reflect ambiguous preferences or subtle perceptual differences.

In the context of video generation, $r_\phi(x)$ serves as a learned reward function that scores videos based on alignment with human preference. The BT/BTT formulations directly correspond to optimizing a discriminative reward model that can distinguish fine-grained perceptual and motion-level differences, and provide a principled objective for preference alignment tasks.


\section{The \modelname Framework}
\label{sec:method}
The framework for our proposed \modelnamefull (\modelname) model consists of training a discriminative model for real and non-real videos using self-supervised contrastive learning leveraging the EBM paradigm, followed by preference-tuning the discriminative model into a reward model that is aligned with human preferences. We use \backbonewithcite as our backbone video generative model owing to its powerful video generation performance and easy reproducibility~\footnote{We are aware of concurrent approaches such as ControlNeXt~\cite{controlnext}. However, they are either not peer-reviewed or not easy to reproduce.}. \cref{fig:pipeline} shows our overall framework.

\begin{figure*}[t]
    \centering
    \begin{subfigure}[b]{0.32\textwidth}
        \centering
        \includegraphics[width=\textwidth]{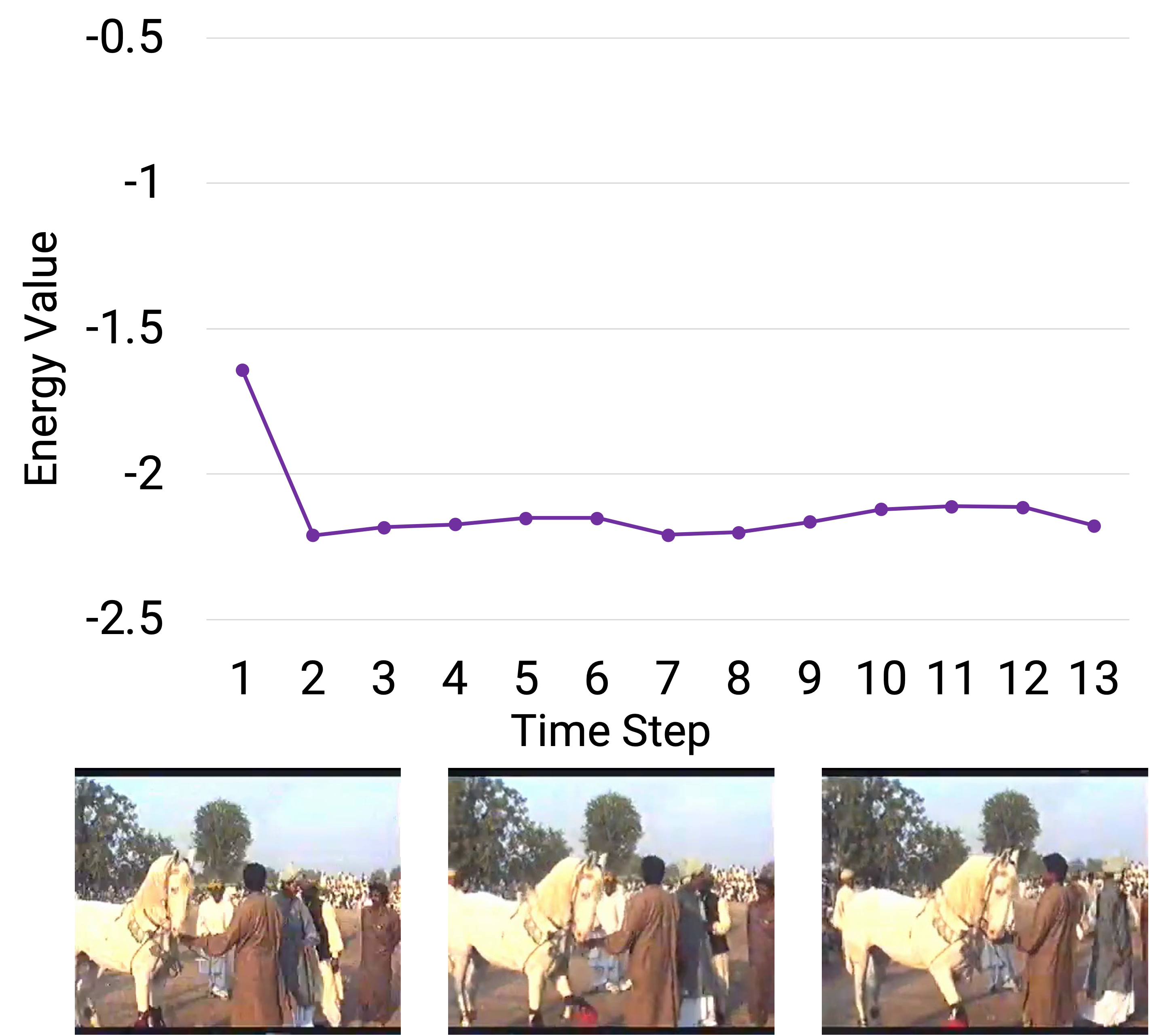}
        \caption{Real video.}
        \label{fig:energy_traj_real}
    \end{subfigure}%
    \hfill
    \begin{subfigure}[b]{0.32\textwidth}
        \centering
        \includegraphics[width=\textwidth]{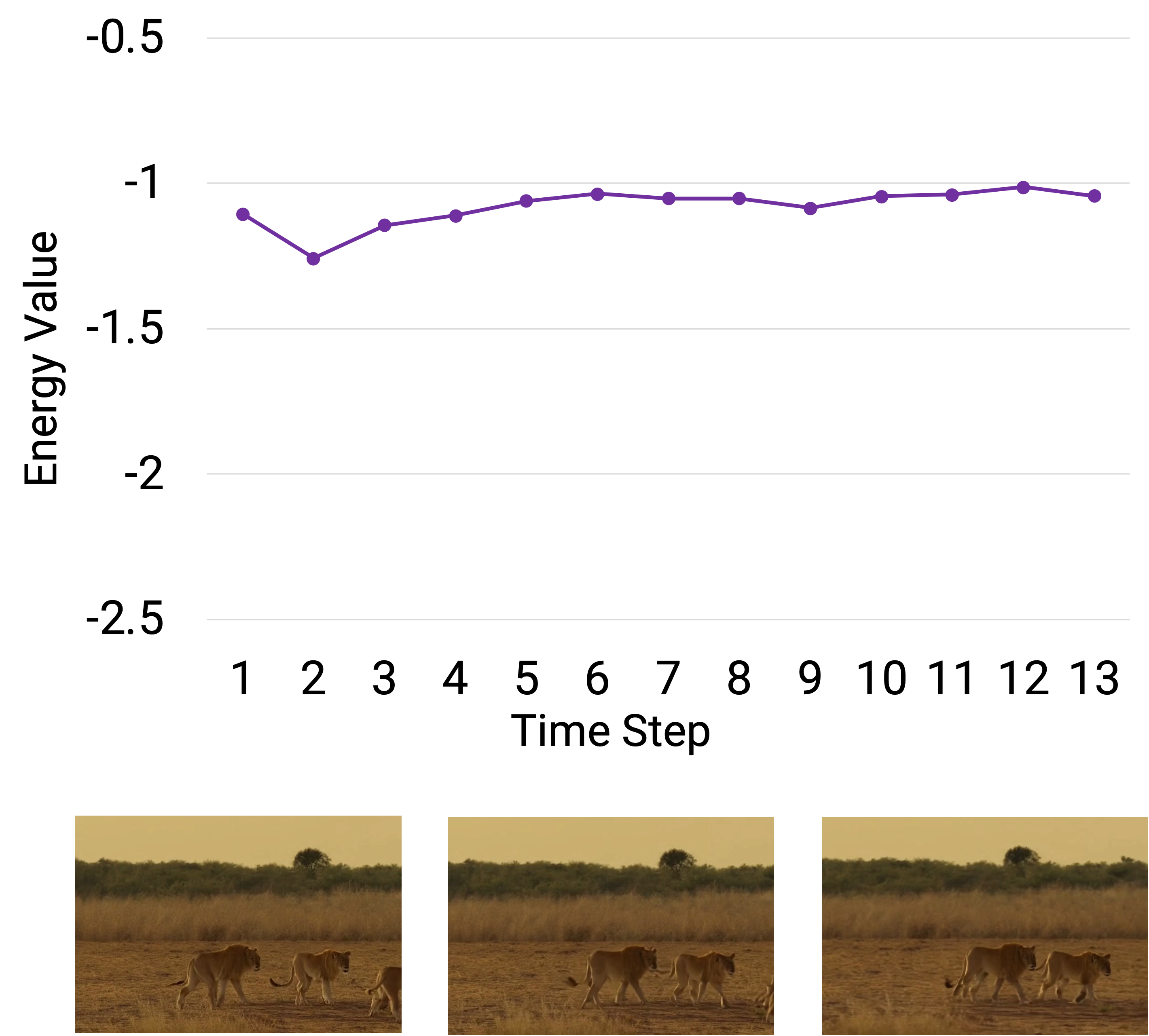}
        \caption{Video generated by \backbonewithcite.}
        \label{fig:energy_traj_cogvideox}
    \end{subfigure}%
    \hfill
    \begin{subfigure}[b]{0.32\textwidth}
        \centering
        \includegraphics[width=\textwidth]{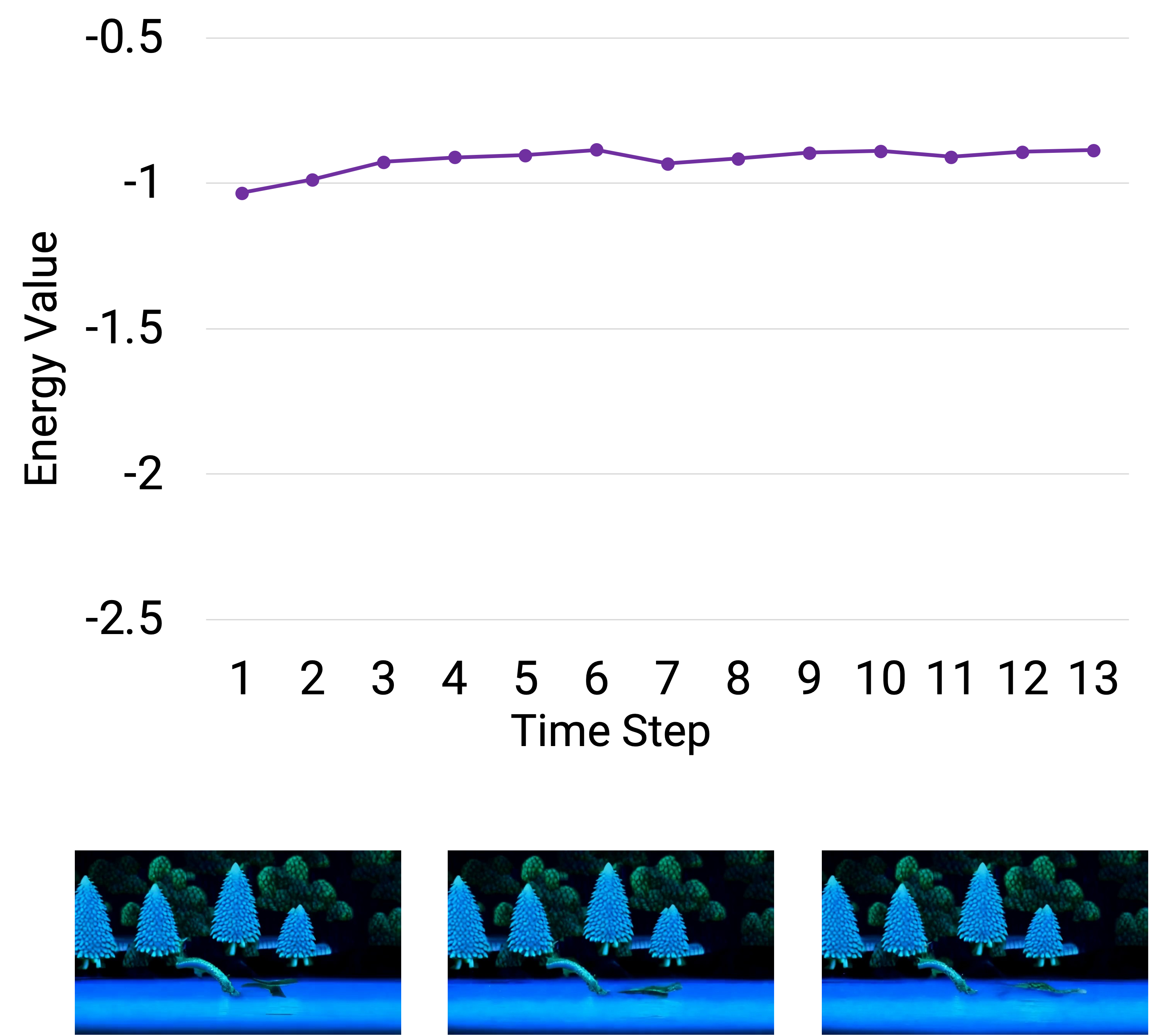}
        \caption{Video generated by OpenSora~\cite{opensora}.}
        \label{fig:energy_traj_opensora}
    \end{subfigure}
    \caption{\textbf{Illustration of energy trajectories predicted by our energy-based model.} For the real video in \textit{(a)}, energy trajectory across the time steps is smooth and stable, indicating consistent temporal dynamics. In contrast, for the generated videos in \textit{(b)} and \textit{(c)}, the energy values fluctuate erratically, reflecting spatial and temporal inconsistencies such as implausible scene lighting and motions.}
    \label{fig:energy_traj}
\end{figure*}

\subsection{Training the Video Discriminative Model}

We build our video discriminative model using some of the transformer layers of \backbonewithcite, followed by a lightweight MLP head to aggregate the spatiotemporal features into a single scalar representation per latent time step (\cref{fig:pipeline} \textit{top}). It takes in the latent space representations $z \in \mZ$ of the input videos $x \in \mX$ as given by the pretrained VAE encoder of \backbone. These latent representations reduce pixel-level redundancy and allow the discriminative model to focus on appearance and motion semantics.

To adapt the \backbone architecture efficiently for contrastive learning, we employ {LoRA} adapters~\cite{lora} on only the last third of its transformer layers, enabling low-rank finetuning with minimal additional parameters. We train the network using the contrastive energy-based formulation of \cref{eq:contrast}, which aims to assign lower energy (higher scores) to real videos and higher energy (lower scores) to all non-real videos.

\textbf{Perturbed Videos As Negative Samples.}
We note that only using model-generated videos as negative samples for contrastive learning tends to result in a trivial discrimination task, where the model learns superficial cues exploiting the domain gap between real and generated videos rather than meaningful spatial structure and temporal dynamics. To counter this, we introduce a more challenging contrastive learning strategy by crafting additional negative samples through controlled perturbations of the latent video representations. Specifically, for each video latent representation $z$, we randomly apply one of the five following manipulations designed to subtly distort the original spatiotemporal structure while preserving overall appearance.\\

\noindent \textbf{Frame Shuffle.}
Given a latent representation $z = \braces{z_t}_{t=1}^T$ over $T$ time steps, we sample a subset of indices
$\mI \subset \{1,\dots,T\}$ and apply a random permutation
$\Pi : \mI \to \mI$.
The modified representation $\widetilde{z}$ satisfies $\widetilde{z}_t = z_{\Pi(t)} \forall t \in \mI$.
This disrupts temporal order while preserving per-frame appearance, requiring the model to detect violations of motion continuity and causal progression.

\noindent \textbf{Frame Drop.}
We randomly sample a set of non-contiguous time steps $\mI$ s.t. $\braces{1} \nsubseteq \mI$, and replace each 
$z_t$ for $t \in \mI$ with its neighbor $z_{t-1}$.
Thus, $\widetilde{z}_t = z_{t-1} \forall t \in \mI$. 
This mimics missing or duplicated frames commonly found in low-frame-rate or corrupted videos, encouraging robustness to irregular temporal sampling.

\noindent \textbf{Noisy Segment Injection.}
We inject Gaussian noise to a random temporal slice $\bracks{s, e}$, $1 \leq s < e \leq T$ such that
\begin{equation*}
  \widetilde{z}_t = z_t + \epsilon_t,\quad \epsilon_t \sim \mN(0, \sigma^2 I) \quad \forall t \in \bracks{s,e}.  
\end{equation*}
This localized corruption simulates sensor noise or compression artifacts, pushing the model to maintain global temporal coherence despite localized errors.

\noindent \textbf{Patch Swap.}
We exchange a random spatial region $\Omega$ between two non-overlapping random temporal slices $[t_1, t_1+\tau]$ and $[t_2, t_2+\tau]$, $\tau \leq T/2$, such that $\forall t \in [t_1, t_1+\tau]$,
\begin{equation*}
\widetilde{z}_{t}(x,y) = 
\begin{cases}
z_{t_2 + (t - t_1)}(x,y), & (x,y) \in \Omega, \\
z_t(x,y), & \text{otherwise}.
\end{cases}    
\end{equation*}
This perturbs object trajectories and motion consistency, requiring joint reasoning of spatial and temporal coherence.

\noindent \textbf{Temporal Slice Swap.}
We choose two random non-overlapping temporal slices 
$[t_1, t_1+\tau]$ and $[t_2, t_2+\tau]$, $\tau \leq T/2$, and swap them, such that
\begin{equation*}
\widetilde{z}_t = 
\begin{cases}
z_{t_2 + (t - t_1)}, & t \in [t_1, t_1+\tau],\\
z_{t_1 + (t - t_2)}, & t \in [t_2, t_2+\tau],\\
z_t, & \text{otherwise}.
\end{cases}
\end{equation*}
This disrupts mid- and large-scale temporal flow while preserving frame-level contents, forcing the model to detect implausible long-range dynamics.

Collectively, these perturbations force the discriminative model to accurately capture spatial structures (such as object appearances and positions) and temporal dynamics (such as object permanence and motion consistency) rather than relying on trivial cues and coarse, frame-level differences. Additionally, with probability $p$, we randomly truncate each video pair to varying lengths, preventing overfitting to fixed-duration inputs and promoting stronger generalization in downstream evaluation settings.

\begin{table*}[t]
\centering
\resizebox{\textwidth}{!}{
\begin{tabular}{lccccC{0.1cm}ccC{0.1cm}cc}
\toprule
\textbf{Method} & 
\textbf{Human Dataset} & 
\textbf{Backbone} &
\multicolumn{2}{c}{\textbf{GenAI-Bench~\cite{genai}}} &&
\multicolumn{2}{c}{\textbf{MonteBench~\cite{visionreward}}} &&
\multicolumn{2}{c}{\textbf{VideoReward-Bench~\cite{videoreward}}} \\
\cmidrule{4-5}\cmidrule{7-8}\cmidrule{10-11}
& \textbf{Size} & \textbf{Size} & w/ties & w/o ties && w/ties & w/o ties && w/ties & w/o ties \\
\midrule

VideoScore~\cite{videoscore}      & \underline{37.6K} & \underline{8B} & 49.03 & 71.69 && 49.10 & 54.90 && 41.80 & 50.22 \\
VisionReward~\cite{visionreward}  & 2000K & 19B & 51.56 & 72.41 && \underline{64.00} & 72.10 && 56.77 & 67.59 \\
VideoReward~\cite{videoreward}    & 182K & \textbf{2B} & \underline{49.41} & \underline{72.89} && 54.20 & 62.25 && \textbf{61.26} & \textbf{73.59} \\
\midrule
\modelname    & \textbf{30.4K} & \textbf{2B} & \textbf{64.26} & \textbf{75.15} && \textbf{66.36} & \textbf{77.76} && \underline{57.01} & \underline{68.57} \\
\modelname ($p=0$) & 30.4K & 2B & 49.36 & 62.41 && 62.79 & \underline{76.12} && 55.42 & 65.37 \\
\modelname (No DM) & 30.4K & 2B & 47.62 & 59.12 && 61.82 & 73.59 && 53.14 & 63.21 \\

\bottomrule
\end{tabular}
}
\caption{\textbf{Video evaluation results on multiple video preference benchmark datasets.} 
We report comparable or better performance in alignment accuracy with human preferences on all benchmark datasets. We also report the human-annotated dataset sizes (in number of samples) and backbone sizes (in number of trainable model parameters). We use the evaluation scheme of Deutsch et al.~\cite{tiesmatter}. Bold: \textbf{best}, underline: \underline{second-best}. Higher values are better for all columns.}
\label{tab:main_res}
\end{table*}

\subsection{Reward Model From Discriminative Model}

Given the trained discriminative model, we build the reward model (\cref{fig:pipeline}, \textit{middle and bottom}) that learns to align with human preferences through preference tuning. For the reward model architecture, we keep the trained LoRA adapters from the discriminative model on the same last third of the transformer layers of the \backbone backbone, and remove the discriminative model's MLP head. We replace it with a new MLP head that takes in the latent representation from the transformer layers and outputs preference scores. Following the approach of VideoReward~\cite{videoreward}, we train an aspect-wise score predictor followed by a relative preference predictor. For the aspect-wise predictor (\cref{fig:pipeline}, \textit{middle}), the output comprises a set of scalar values $q \in \bbR^\mQ$, one for each aspect of video quality that requires alignment with human preferences, \textit{e.g.}, realism, smoothness, temporal coherence, and color consistency.
We consider numeric scores for human preferences, such as the commonly-used 1 to 5 Likert scale values, and train the reward model end-to-end using a Mean Squared Error (MSE) loss to regress to the human scores.

Once the aspect-wise score predictor is trained, we introduce a lightweight linear head to map the aspect-wise scores to a scalar score for relative preferences (\cref{fig:pipeline}, \textit{bottom}), and train it on pairwise human preferences under the Bradley-Terry loss (\cref{eq:bt,eq:btt}). This two-step procedure yields a stable, interpretable reward model that accurately reflects human perceptual preferences and is suitable for downstream RL and evaluation tasks.


\section{Experiments and Results}
\label{sec:exp}

We provide the implementation details for both the discriminative and reward models under \modelname and analyze the performance of these models and relevant ablations on large-scale benchmark datasets for video preference.

\subsection{Implementation Details}

\textbf{Discriminative Model.}
For all experiments of the discriminative model, we fix the LoRA rank $r=8$ and scaling factor $\alpha=8$, providing a reasonable trade-off between parameter efficiency and representational capacity.
The pretrained and frozen \backbone autoencoder compresses input videos of $4f + 1$ frames into $f + 1$ latent frames. Because the \backbone autoencoder is causal, the latent representation at each time step depends on all preceding latents, making the resulting energy sequences interpretable as a measure of temporal perplexity.

Training follows a contrastive energy-based objective $\mL_{\text{contrast}}$ (\cref{eq:contrast}) with $\beta=0.2$, where real videos are the positive samples and generated and perturbed videos are the negative samples. Our corpus contains approximately 20K real Internet-sourced videos (each cropped into 6-second segments) and 30K videos generated using various video generation models, including \textit{Gen3}~\cite{Runway2024Gen3}, \textit{Luma Dream Machine}~\cite{LumaLabs2024DreamMachine}, \textit{\backbone}~\cite{cogvideox}, \textit{OpenSora}~\cite{opensora}, and \textit{VideoCrafter2}~\cite{videocrafter2}. Prompts are drawn from publicly available datasets such as \textit{VisionReward}~\cite{visionreward} and \textit{VideoReward}~\cite{videoreward}. These 50K videos require no human annotation, synthetic negatives are generated
  via perturbations~(\cref{sec:method}). Only reward modeling uses
  human-annotated data (30.4K videos). Further, we note that most current video preference datasets consist of video clips of uniform length, commonly 6 seconds. Models trained from scratch on such data often develop an implicit length bias, resulting in poor performance on videos of varying lengths. We mitigate this issue by training the discriminative model on videos of variable lengths, randomly slicing clips between 2 and 6 seconds and selecting clips shorter than 6 seconds with probability $p=0.25$. This exposes the model to a broader temporal distribution, enabling the learning of duration-invariant representations.


As illustrated in \cref{fig:energy_traj}, real videos (\cref{fig:energy_traj_real}) maintain smooth, low-variance energy curves consistent with natural temporal continuity. Conversely, synthetic or perturbed videos (\cref{fig:energy_traj_cogvideox,fig:energy_traj_opensora}) exhibit abrupt fluctuations, indicating irregular temporal transitions. This successfully aligns with our contrastive objective: assigning lower energy to temporally coherent sequences and higher energy to incoherent or unpredictable motion patterns. 

\begin{figure*}[t]
    \centering
    \begin{subfigure}[t]{0.32\textwidth}
        \centering
        \includegraphics[width=\textwidth]{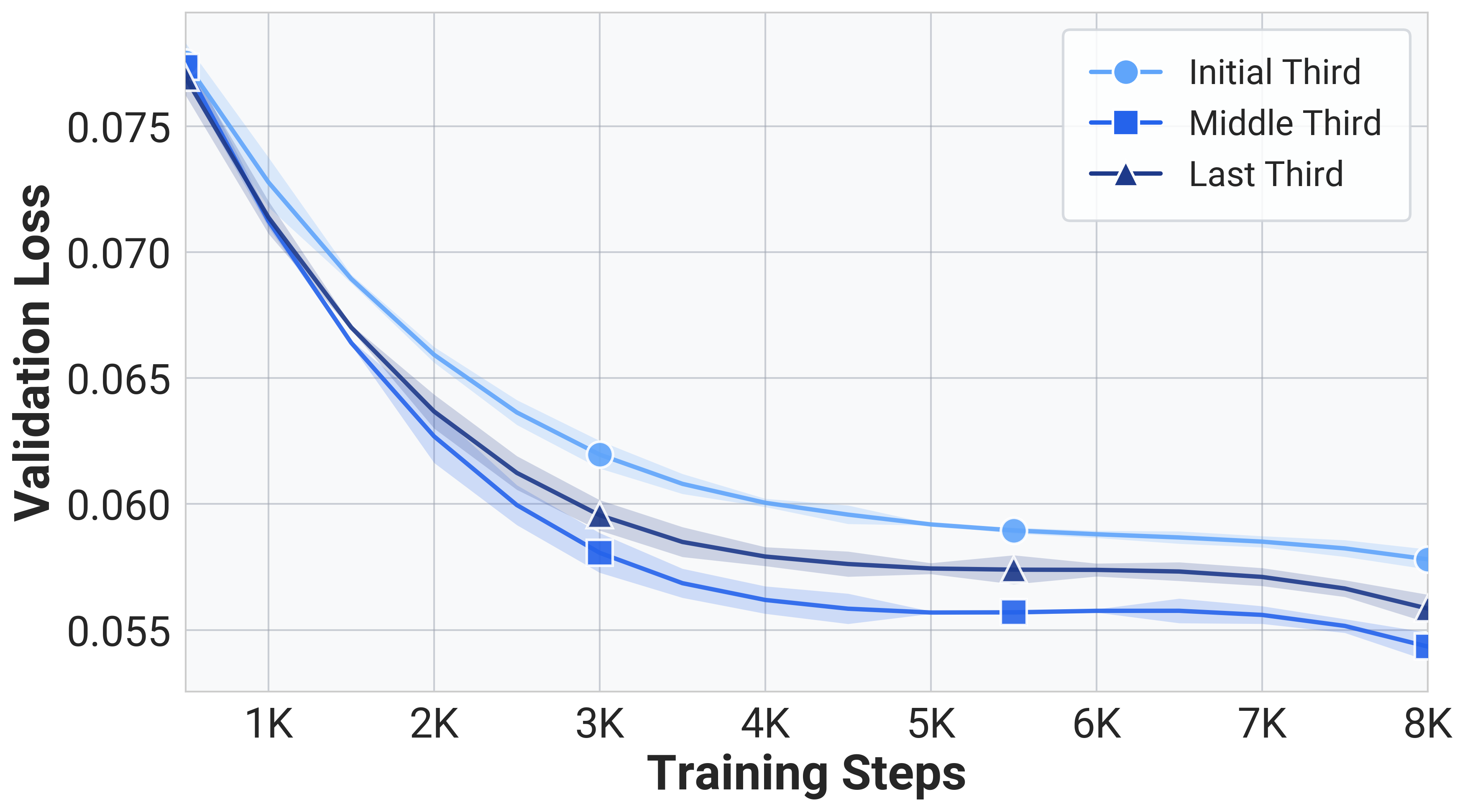}
        \caption{\textbf{Validation losses across LoRA placements.} Applying LoRA to the middle third of the transformer achieves the lowest validation losses for aspect-wise score prediction of the reward model, indicating that mid-level representations are most amenable to efficient adaptation.}
        \label{fig:lora_loc_val_loss}
    \end{subfigure}%
    \hfill
    \begin{subfigure}[t]{0.32\textwidth}
        \centering
        \includegraphics[width=\textwidth]{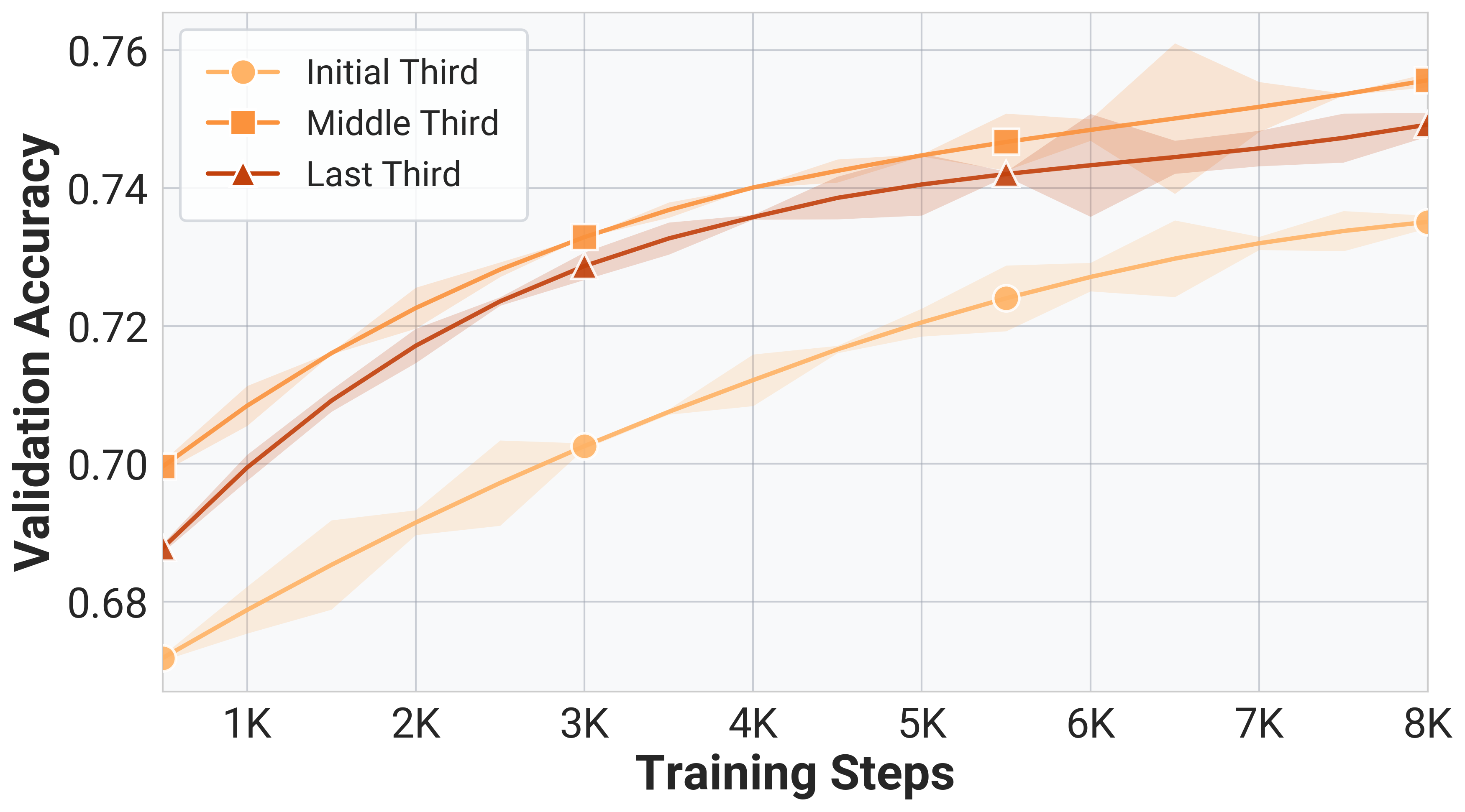}
        \caption{\textbf{Validation accuracies across LoRA placements.} Applying LoRA to the middle third of the transformer also achieves the highest validation accuracies for relative preference predictions of the reward model.}
        \label{fig:lora_loc_val_acc}
    \end{subfigure}%
    \hfill
    \begin{subfigure}[t]{0.32\textwidth}
        \centering
        \includegraphics[width=\textwidth]{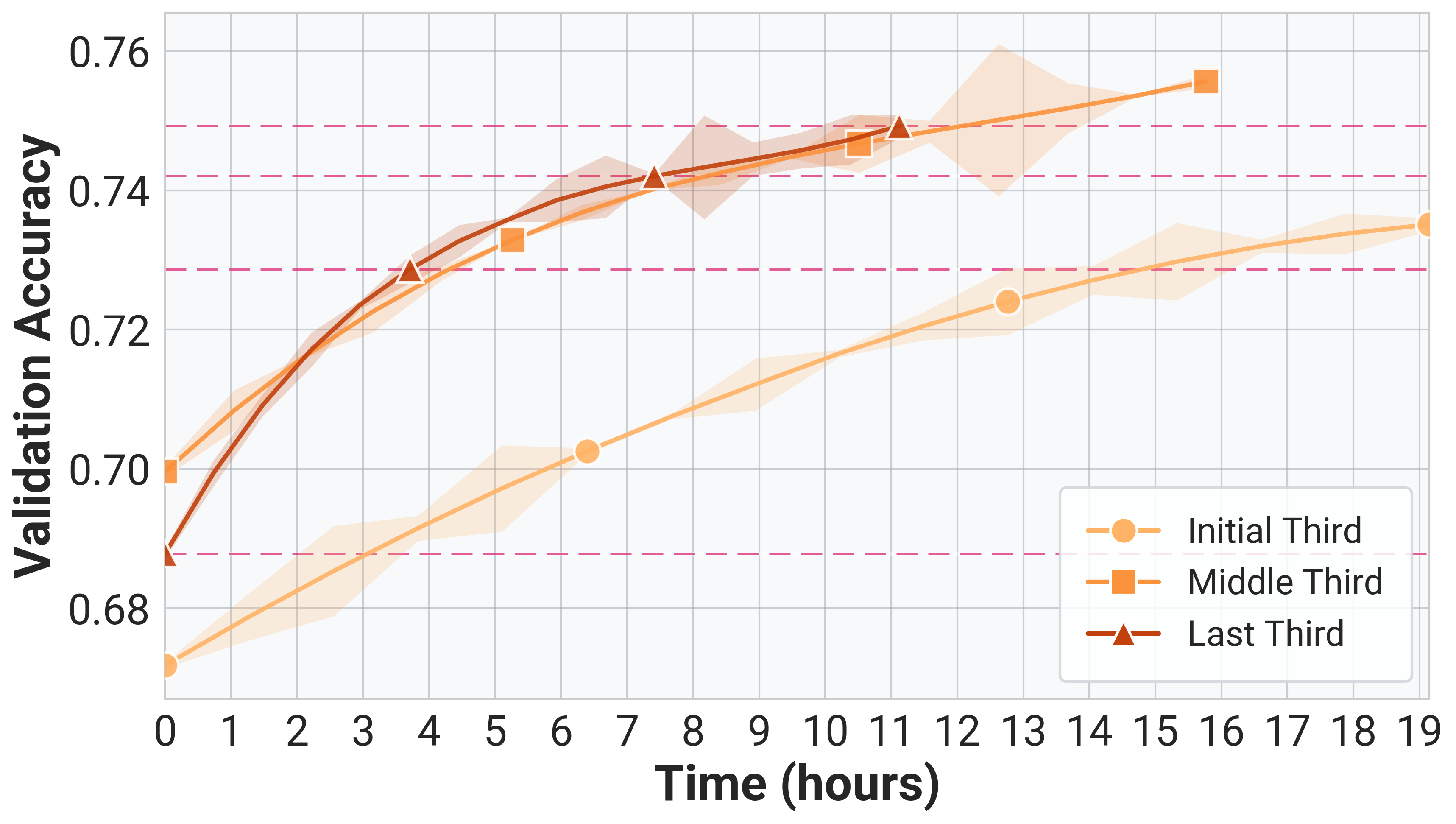}
        \caption{\textbf{Validation accuracy vs time.} While all configurations eventually converge, applying LoRA to later layers leads to faster convergence but slightly worse generalization, highlighting a trade-off between efficiency and representation depth.}
        \label{fig:lora_loc_val_acc_v_time}
    \end{subfigure}
    \caption{\textbf{Effect of LoRA placement within the backbone transformer.} We compare applying LoRA to the initial third, middle third, and last third of the transformer layers. The middle-layer configuration achieves the best overall performance, while the last-layer configuration provides faster training with minimal loss in accuracy.}
    \label{fig:lora_ablations}
\end{figure*}


\noindent\textbf{Reward Model.}
To train our reward model, we use $\mQ = 21$ video quality aspects corresponding to the 21 annotated attributes in the \textit{VisionReward}~\cite{visionreward} dataset.
We train this reward model end-to-end using the \texttt{train} split of \textit{VisionReward} and the mean squared error loss, $\mL_{\text{MSE}}$. Subsequently, an aggregation head maps the 21 predicted attributes to a single scalar reward score, which is then fine-tuned using preference supervision on the \texttt{regression} split.
All reward-model experiments use LoRA adapters (rank $r=8$, $\alpha=8$) to enable parameter-efficient fine-tuning consistent with our discriminative setup. During pairwise preference training, we also truncate videos with probability $p=0.2$, assuming that the underlying preference is preserved after truncation. When truncation is applied, the video length is randomly reduced to a duration between 2 and 6 seconds. 

\subsection{Benchmarks and Baselines}
We evaluate \modelname across three state-of-the-art video preference benchmarks: \textit{GenAI-Bench}~\cite{genai}, \textit{MonteBench}~\cite{visionreward}, and \textit{VideoReward-Bench}~\cite{videoreward}, alongside their corresponding baseline models.
\modelname demonstrates strong generalization across all three datasets (\cref{tab:main_res}). On GenAI-Bench, it surpasses previous baselines by $24.63\%$ and $3.1\%$ under the tie and no-tie settings, respectively. For MonteBench, we observe gains of $3.68\%$ and $7.85\%$ on the same evaluation metrics. On VideoReward-Bench, \modelname ranks second, trailing the VideoReward model by roughly $5\%$. We attribute this gap to a distribution shift between VideoReward-Bench and the other benchmarks. The absence of public training data for this benchmark further limits direct alignment with its preference domain, making its evaluation setting uniquely challenging.

\subsection{Ablations}
We perform several ablations to analyze the impacts of our proposed model components and training setups.

\noindent\textbf{Effect of LoRA Placement.}
We study the impact of applying LoRA adapters at different depths within the transformer architecture. While it is commonly assumed that adding LoRA to the final few layers is most effective, since these layers typically capture fine-grained details and task-specific semantics, we observe some caveats.

Our \backbone backbone contains 30 Transformer layers, which we group into three segments: the initial 10 layers (\emph{initial third}), the middle 10 layers (\emph{middle third}), and the final 10 layers (\emph{last third}). As illustrated in \cref{fig:lora_ablations}, inserting LoRA adapters into the \emph{middle third} yields the best results, achieving both the lowest validation loss and highest downstream accuracy. In comparison, adapting the last third layers performs slightly worse, indicating that mid-level features are particularly important for effective model adaptation.

Nevertheless, for our proposed model, we choose to apply LoRA to the \textit{last third} of the transformer layers. As observed in \cref{fig:lora_loc_val_acc_v_time}, this configuration offers approximately a $1.5\times$ faster training speed while incurring only a marginal decrease in validation performance. Thus, it provides a practical trade-off between computational efficiency and performance.

\begin{figure}[t]
    \centering
    \begin{subfigure}[b]{0.48\textwidth}
        \centering
        \includegraphics[width=0.8\textwidth]{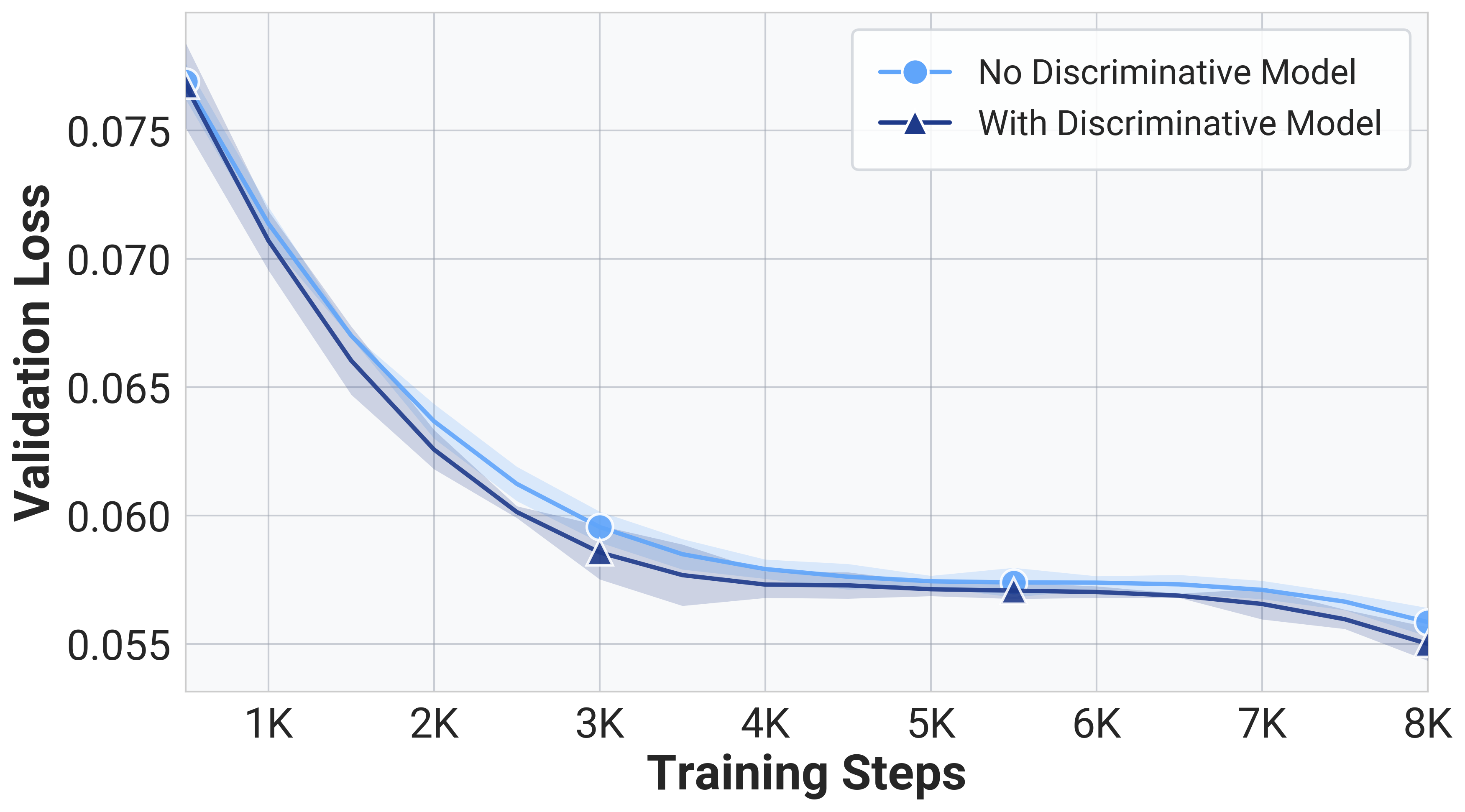}
        \caption{Validation losses for aspect-wise predictor training.}
        \label{fig:dm_val_loss}
    \end{subfigure}
    \hfill
    \begin{subfigure}[b]{0.48\textwidth}
        \centering
        \includegraphics[width=0.8\textwidth]{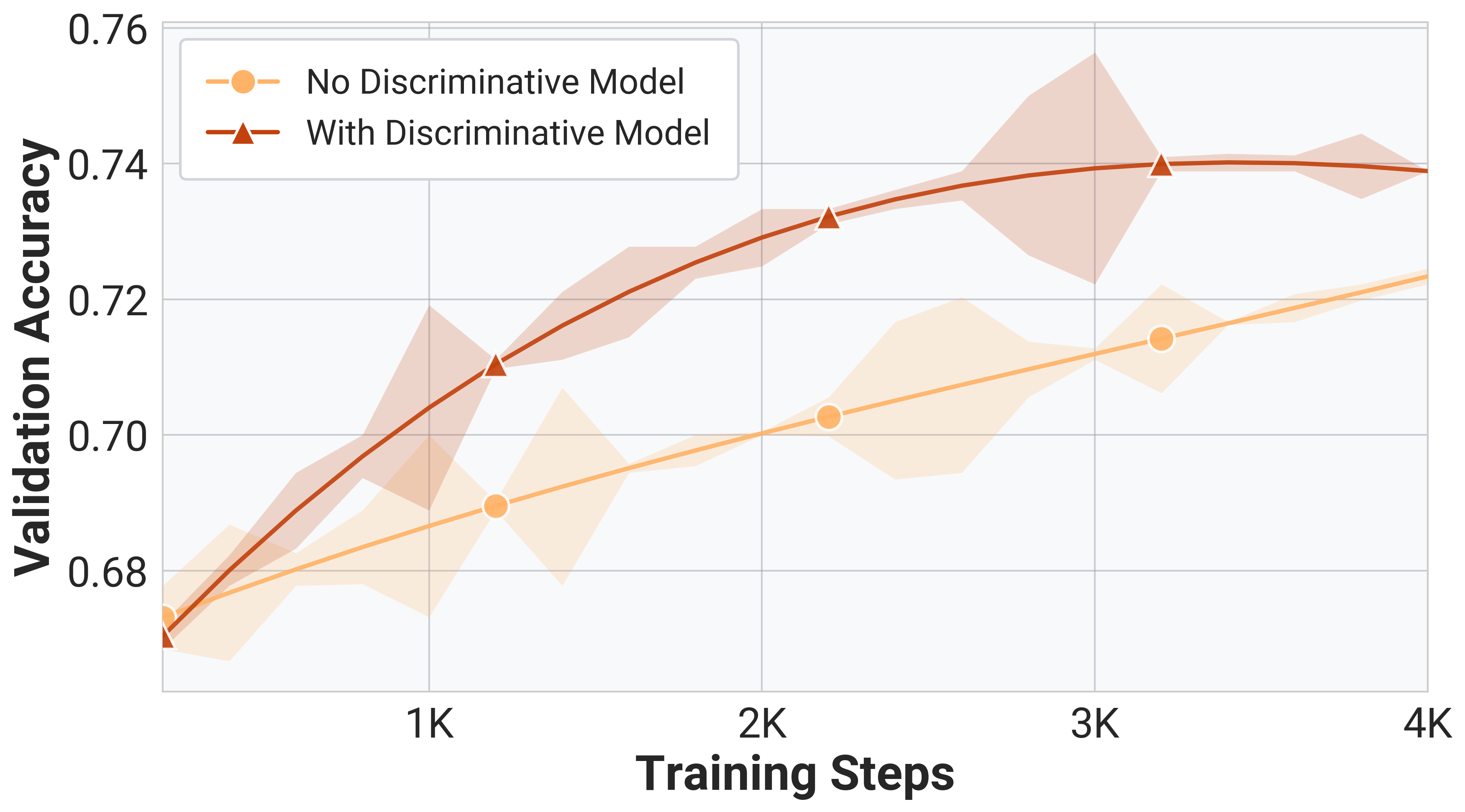}
        \caption{Validation accuracies for relative preference training.}
        \label{fig:dm_val_acc}
    \end{subfigure}
    \caption{\textbf{Effect of the discriminative model.} Initializing the reward model with the trained discriminative model leads to lower validation losses and higher validation accuracies throughout training.}
    \label{fig:dm_ablations}
\end{figure}

\noindent\textbf{Effect of the Discriminative Model.}
We analyze the impact of the discriminative model on the performance of the reward model. Comparing \modelname and \modelname (no DM) in \cref{tab:main_res}, we observe that the discriminative model leads to improvements of about $5\%$ in the alignment accuracy. We further illustrate the improvements in \cref{fig:dm_val_loss,fig:dm_val_acc}.
\cref{fig:dm_val_loss} shows that using the discriminative model leads to lower validation losses and higher validation accuracies throughout training the aspect-wise predictor and the relative preferences, respectively, for the reward model.
Our findings are consistent with insights from \cite{repa}, which suggest that discriminative learning transforms generative representations into features better aligned for downstream discrimination tasks. Initializing the reward model with trained discriminative features provides a stronger inductive bias for learning effective reward predictors.

\noindent\textbf{Effect of Perturbed Videos as Negative Samples.}
To better understand the role of perturbed videos as negative samples, we perform an ablation study comparing \modelname trained under two different regimes: \textit{(i)} using real videos as positives and generated videos as negatives, and \textit{(ii)} using our proposed perturbed videos as additional negatives alongside real and generated samples. 

\begin{figure}[t]
    \centering
    \begin{subfigure}[b]{0.48\textwidth}
        \centering
        \includegraphics[width=0.8\textwidth]{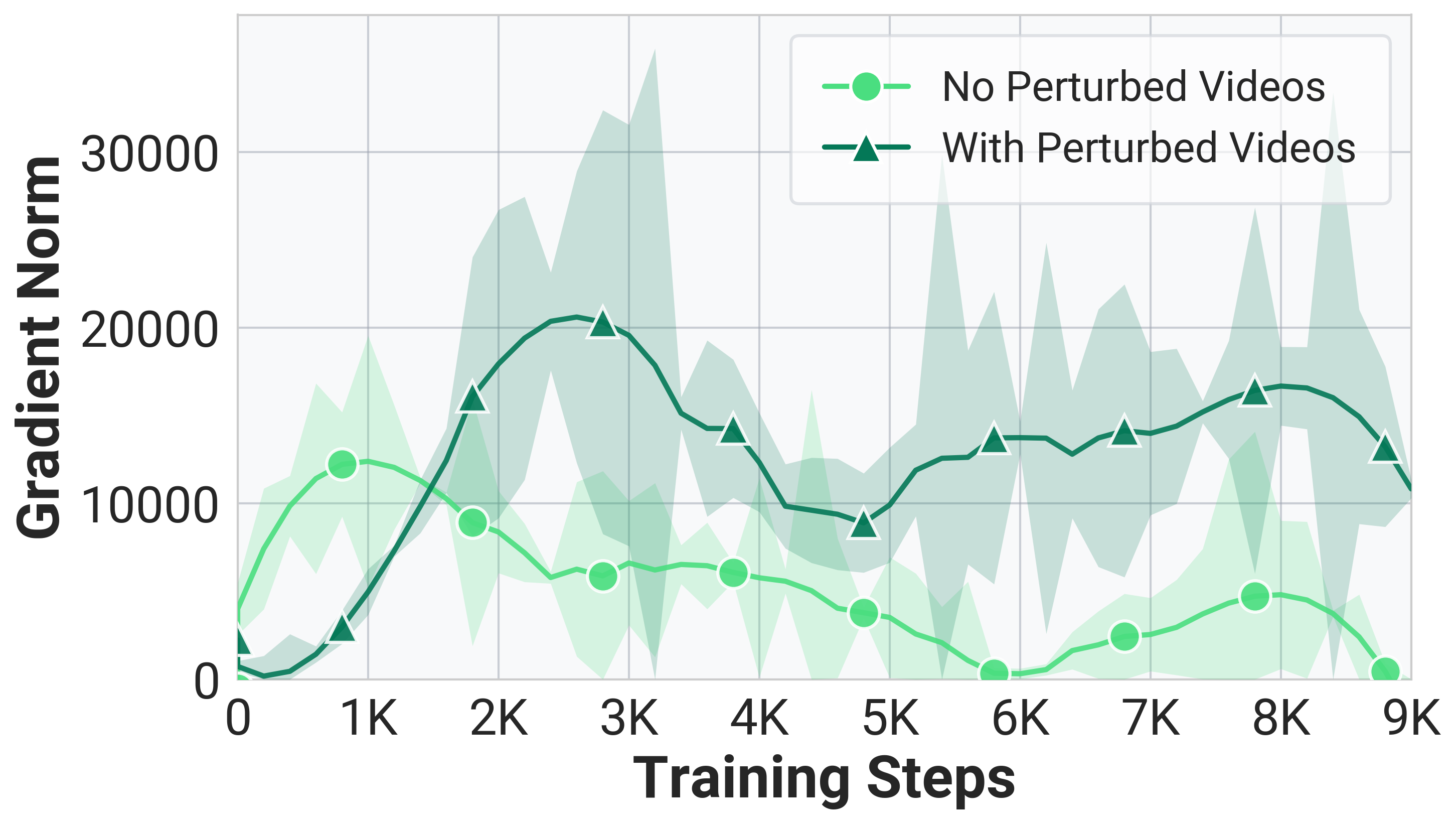}
        \caption{Average gradient norm during training for different data regimes.}
        \label{fig:perturbed_videos_grad_norm}
    \end{subfigure}
    \hfill
    \begin{subfigure}[b]{0.48\textwidth}
        \centering
        \includegraphics[width=0.8\textwidth]{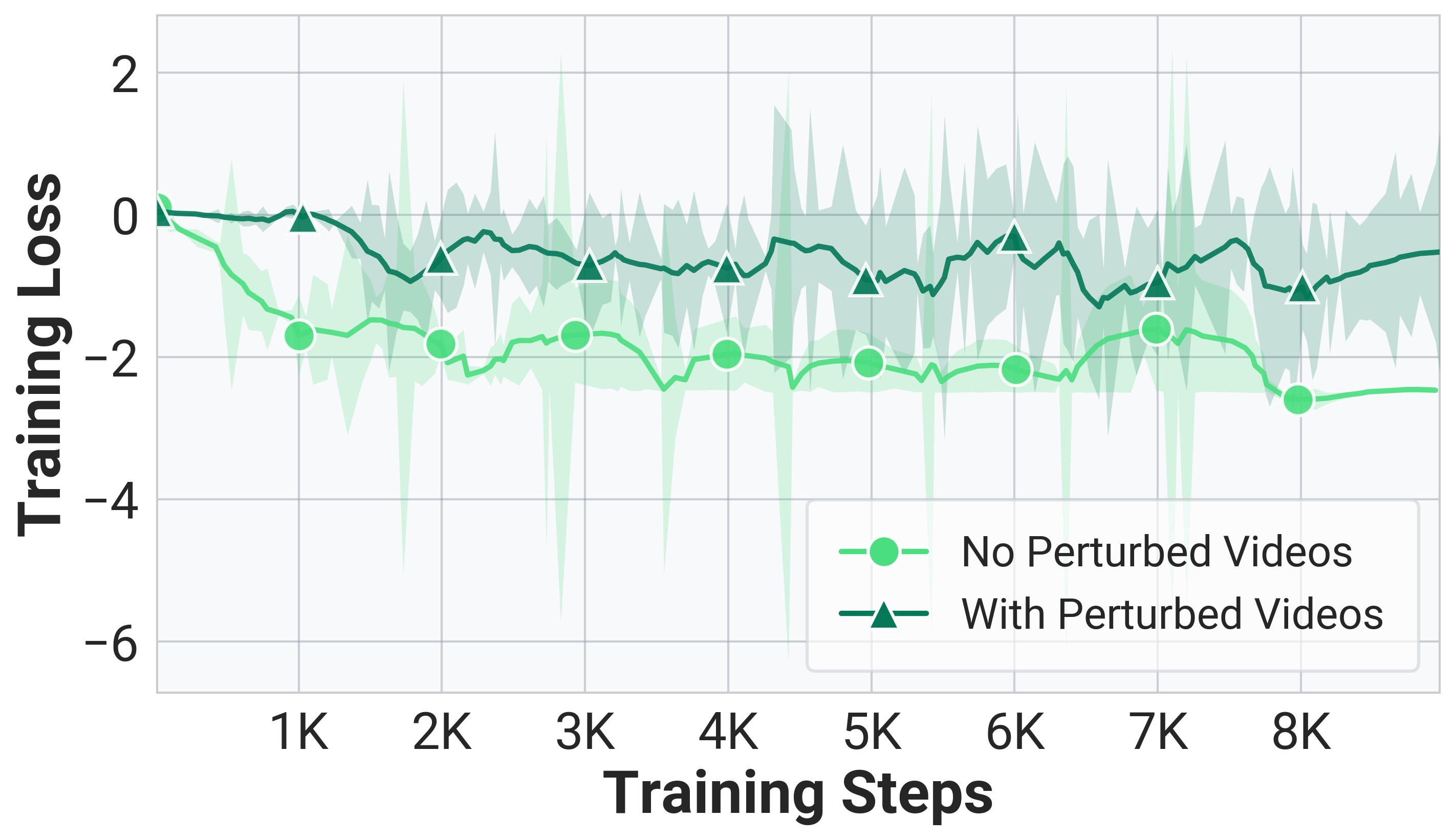}
        \caption{Training loss curves under different data regimes.}
        \label{fig:perturbed_videos_train_loss}
    \end{subfigure}
    \caption{\textbf{Effect of perturbed videos as negative samples.} Our discriminative model, when trained only on real and generated videos, exhibits early saturation and vanishing gradients. In contrast, when perturbed videos are augmented as negative samples in training, the model maintains informative gradients and achieves smoother convergence.}
    \label{fig:perturbed_videos_ablations}
\end{figure}

As shown in \cref{fig:perturbed_videos_ablations}, our discriminative model trained solely on real and generated videos quickly converges to trivial solutions. The losses in these settings rapidly approach the soft lower bound imposed by the regularization constant (\cref{fig:perturbed_videos_train_loss}), indicating that the model saturates early and fails to capture meaningful discriminative cues. This is also reflected in the average gradient norm, which initially spikes but then rapidly diminishes to near-zero values, suggesting a cessation of effective learning.

In contrast, the introduction of perturbed videos as negatives substantially improves the learning dynamics. The gradient norms remain higher and more stable over training, implying that the model continues to refine its energy landscape. Further, the loss curve for the synthetic-augmented model decreases more gradually, demonstrating that the model benefits from a more challenging and informative contrastive objective. These results collectively indicate that synthetic perturbations enhance the model's ability to learn fine-grained spatiotemporal features.

\noindent\textbf{Effect of Variable Video Lengths in Training.}
As shown in \cref{tab:main_res}, \modelname (which has $p=0.25$) achieves $1-15\%$ better alignment accuracy compared to \modelname $(p=0)$, which is trained only with uniform-length video clips. Since \modelname $(p=0)$ never observes very different-length (\textit{e.g.}, 2-second) clips during training, it fails to generalize to the diverse temporal distributions in video preferences.


\section{Conclusion}
\label{sec:conclusion}

We have presented \modelname, a data-efficient, self-supervised reward modeling framework that matches and surpasses alignment accuracy with human preferences compared to state-of-the-art VLM-based baselines on diverse benchmark datasets for video preference. \modelname derives its performance primarily from our proposed discriminative model, which is trained to understand fine-grained spatial distortions and motion artifacts, and is then utilized to train our reward model. We further demonstrate the benefits of including our proposed method for crafting negative samples for the discriminative model through controlled perturbations of real videos, which improves gradient stability, prevents early convergence, and yields more reliable optimization behavior. However, unlike VLM-based evaluators, \modelname does not inherently provide a conversational interface, which remains a limitation for interaction-heavy evaluation scenarios. We also aim to extend this framework to multi-modal reward alignment settings and explore the impacts of incorporating prompts for video generation in the discriminative and reward models.

\section{Acknowledgment}
This research used both Delta (NSF award OAC 2005572) and DeltaAI (NSF
award OAC 2320345) advanced computing systems, and computing resources
provided by Illinois Computes and NAIRR Pilot NAIRR250157.

\small
{
    \bibliographystyle{ieeenat_fullname}
    \bibliography{main}
}

\clearpage
\setcounter{page}{1}
\setcounter{section}{0}
\renewcommand\thesection{\Alph{section}}
\counterwithin{equation}{section}
\counterwithin{figure}{section}
\counterwithin{table}{section}
\maketitlesupplementary

\section{Perturbing Real Videos}

\begin{table}[t]
\centering
\resizebox{\columnwidth}{!}{
\begin{tabular}{lcc}
\toprule
\textbf{Perturbation Type} & \multicolumn{2}{c}{\textbf{Sampling Probability}} \\
& \textbf{Real Video} & \textbf{Generated Video} \\
\midrule
Temporal Slice Swap  & $0.04$ & $0.05$ \\
Noisy Segments       & $0.10$ & $0.15$ \\
Frame Shuffle        & $0.18$ & $0.25$ \\
Frame Drop           & $0.18$ & $0.25$ \\
Patch Swap           & $0.20$ & $0.30$ \\
\bottomrule
\end{tabular}
}
\caption{\textbf{Sampling probabilities assigned to each perturbation type based on gradient-based difficulty analysis.} Note that a fixed probability of $0.3$ is reserved for contrasting real videos with model-generated videos, and hence the probabilities of perturbations for real videos sum to $0.7$.}
\label{tab:perturb_probs}
\end{table}

To explicitly train our reward model to recognize subtle temporal and spatial inconsistencies in videos, we introduce a diverse set of perturbations that simulate realistic failure modes observed in video generative models. These perturbations are designed to degrade semantic coherence, temporal smoothness or visual fidelity while preserving overall structure -- thereby creating hard negative samples that encourage the model to learn fine-grained consistency cues rather than relying on trivial artifacts.

For training, each real video is paired with a randomly chosen perturbed counterpart produced through one of five operators: frame shuffle, frame drop, noisy segments, patch swap, or temporal slice swap. We show representative examples of these perturbations in the supplementary video \texttt{supp\_video\_perturbations.mp4}.








\begin{figure}[t]
    \centering
    \begin{subfigure}[b]{0.48\textwidth}
        \centering
        \includegraphics[width=\textwidth]{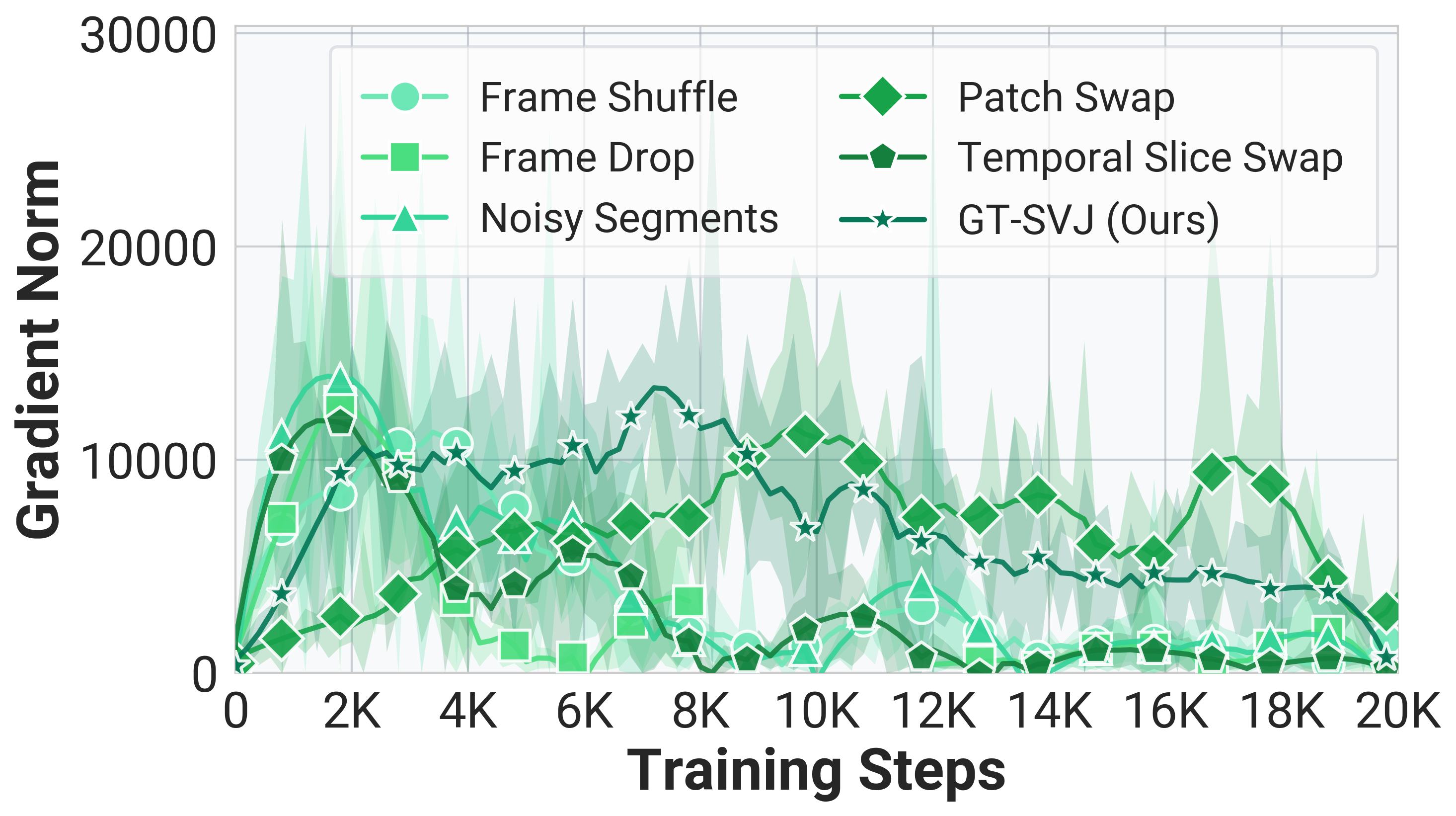}
        \caption{Average gradient norm during training from using each perturbation type individually.}
        \label{fig:perturb_abl_grad_norm}
    \end{subfigure}
    \hfill
    \begin{subfigure}[b]{0.48\textwidth}
        \centering
        \includegraphics[width=\textwidth]{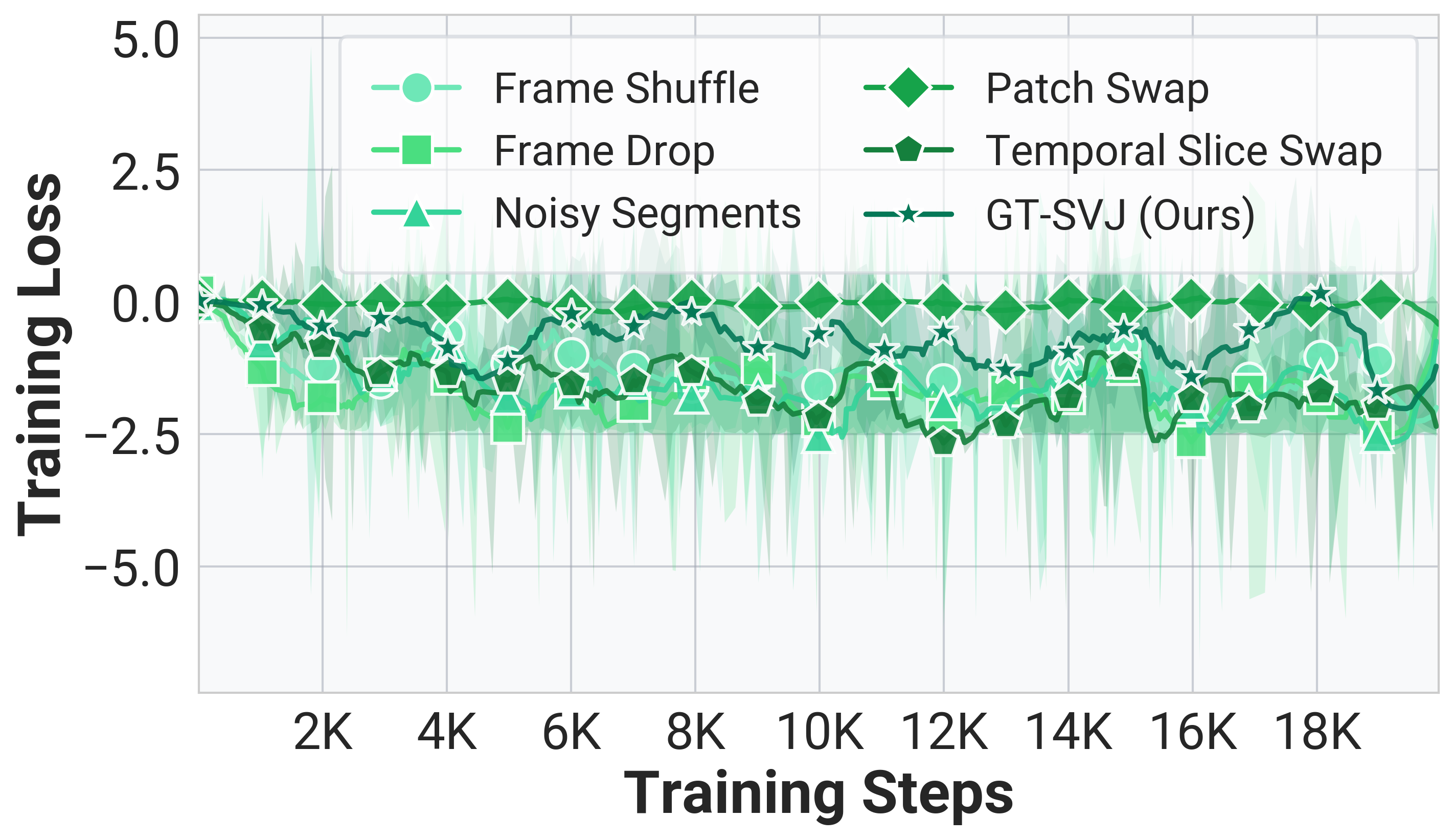}
        \caption{Training loss curves from using each perturbation type individually.}
        \label{fig:perturb_abl_train_loss}
    \end{subfigure}
    \caption{\textbf{Effect of using each perturbation type individually to generate negative samples.} We show the ease or difficulty of training the discriminative model when using each perturbation type individually.}
    \label{fig:perturbation_ablations}
\end{figure}

\subsection{Adaptive Perturbation Sampling}

At each training step, rather than uniformly sampling negatives from all five types of perturbations to augment our training dataset, we assign a fixed probability of selection  based on a perturbation's \textit{relative difficulty}, defined by how easily the discriminative model can learn to distinguish the perturbed sample in isolation. Difficulty is quantified through the magnitude of the gradient of the loss with respect to the model parameters when trained on a single perturbation type. Perturbations that induce lower initial gradients (\textit{i.e.}, are harder to detect) are sampled with higher probability to ensure sufficient exposure during training, while easier perturbations receive lower weights to avoid gradient domination early on (See \cref{fig:perturbation_ablations}).

Formally, let $\mL_p$ denote the loss corresponding to perturbation type $p$, and let $\nabla_\theta \mL_p$ be the gradient with respect to model parameters $\theta$. Then we obtain the sampling probability $\mP$ for perturbation $p$ based on
\begin{equation}
    \mP\parens{p} \propto \frac{1}{\modulus{\nabla_\theta \mL_p}}.
\end{equation}

In practice, we track the gradients for each perturbation type across independent training runs and quantify its effective difficulty based on the rate at which its gradient norm decays. Specifically, we monitor $\modulus{\nabla_\theta \mL_p\parens{t}}$ over optimization steps and apply early stopping when the gradient norm falls below a small threshold $\epsilon=1500$, indicating saturation. The decay time
\begin{equation}
\tau_p = \min \{ t : \modulus{\nabla_\theta \mL_p(t)} \leq \epsilon \}
\end{equation}
serves as a proxy for perturbation difficulty, where smaller values correspond to easier perturbations that are rapidly learned in isolation.

We employ an inverse decay-based heuristic to determine perturbation sampling probabilities, assigning higher weights to distortions whose gradients decay more slowly over training. This design choice prevents harder-to-learn perturbations from being under-represented and ensures sustained exposure to challenging negative samples. After normalization, the resulting probabilities are fixed for the remainder of training to enable stability and reproducibility across runs.
We report the empirically calibrated sampling distributions used for all experiments in \cref{tab:perturb_probs}.

Although all perturbation types exhibit comparable peak gradient magnitudes during early training, we observe substantial variation in their gradient decay dynamics. Perturbations whose gradients vanish rapidly contribute limited long-term learning signals and are consequently down-weighted. In contrast, perturbations that maintain persistent gradients are prioritized, enforcing continuation of optimization and improved discriminative robustness.

\begin{figure}
    \centering
    \includegraphics[width=\columnwidth]{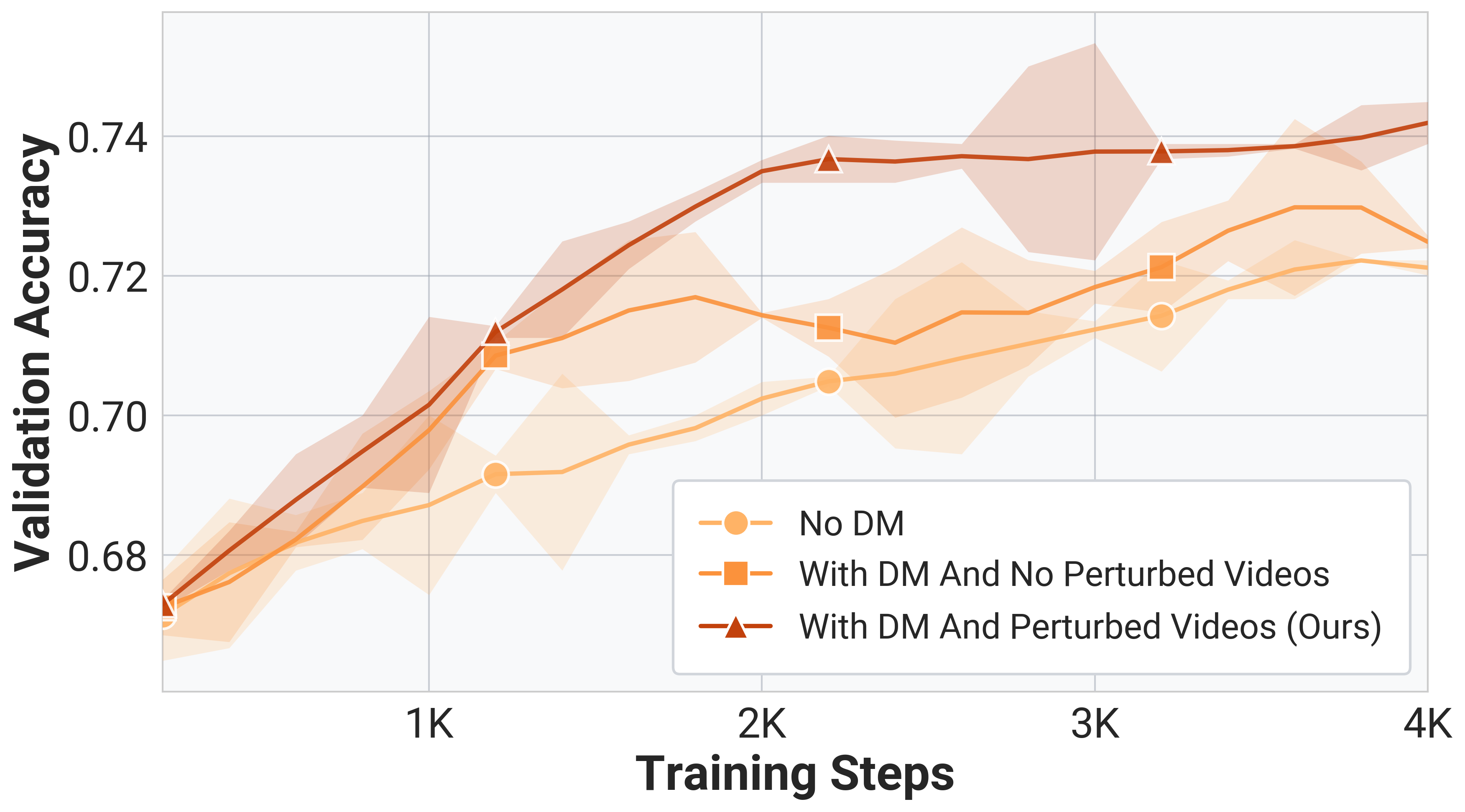}
    \caption{\textbf{Effect of using all types of perturbed videos in training the discriminative model (DM).} We observe significantly faster convergence to 2-3\% better validation alignment accuracies to human preferences.}
    \label{fig:perturb_val_acc}
\end{figure}

\subsection{Impact on Learning}

These perturbations collectively encourage the model to develop sensitivity to both spatial realism and temporal coherence. As shown in \cref{fig:perturbation_ablations}, perturbation types that produce visually plausible yet semantically inconsistent videos lead to smoother loss curves but smaller gradients, justifying their greater difficulty and higher sampling frequency for data augmentation. This design effectively acts as a curriculum, guiding the model from easy-to-detect artifacts toward more nuanced inconsistencies.

We also show the collective benefits of using all the perturbations in training our discriminative model in \cref{fig:perturb_val_acc}. Adding perturbed videos to training consistently leads to faster convergence to 2-3\% better alignment accuracies with human preferences during inference.

\section{Qualitative Results}

We show qualitative results of \modelname and the different baselines on video preference in \texttt{supp\_video\_results.mp4}. We show the performance on human preference alignment for each method on five randomly selected video pairs from each of the three benchmark datasets, GenAI-Bench~\cite{genai}, MonteBench~\cite{visionreward}, and VideoReward-Bench~\cite{videoreward}.

\end{document}